\documentclass[a4paper]{article}

\usepackage{INTERSPEECH2018}
\usepackage{multirow}

\title{A COMPARABLE STUDY OF MODELING UNITS FOR END-TO-END MANDARIN SPEECH RECOGNITION}
\name{Wei Zou, Dongwei Jiang, Shuaijiang Zhao, Xiangang Li}
\address{
  AI Labs, Didi Chuxing, Beijing, China
  }
\email{\{zouwei,jiangdongwei,zhaoshuaijiang,lixiangang\}@didichuxing.com}

\begin{document}

\maketitle

\begin{abstract}
  End-To-End speech recognition have become increasingly popular in mandarin speech recognition and achieved delightful performance.
  Mandarin is a tonal language which is different from English and requires special treatment for the acoustic modeling units. There have been several different kinds of modeling units for mandarin such as phoneme, syllable and Chinese character.
  In this work, we explore two major end-to-end models: connectionist temporal classification (CTC)  model and attention based encoder-decoder model for mandarin speech recognition. We compare the performance of three different scaled modeling units: context dependent phoneme(CDP), syllable with tone and Chinese character. 
  We find that all types of modeling units can achieve approximate character error rate (CER) in CTC model and the performance of Chinese character attention model is better than syllable attention model. Furthermore, we find that Chinese character is a reasonable unit for mandarin speech recognition. On DidiCallcenter task, Chinese character attention model achieves a CER of 5.68\% and CTC model gets a CER of 7.29\%,  on the other DidiReading task, CER are 4.89\% and 5.79\%, respectively. Moreover, attention model achieves a better performance than CTC model on both datasets.
\end{abstract}

\section{Introduction}
Traditional speech recognition includes separate modeling components, including acoustic, phonetic and language models.
These components of the system are trained separately, thus each component’s errors would extend during the process.
Besides, building the components requires expert knowledge, for example, building a language model requires linguistic knowledge.
The acoustic model is used to recognize context-dependent (CD) states or phonemes \cite{deng2013recent,schmidhuber2015deep}, by bootstrapping from an existing model which is used for alignment. 
The pronunciation model maps the phonemes sequences into word sequences, then the language model scores the word sequences.
A weighted finite state transducer (WFST) \cite{miao2015eesen} integrates these models and do the decoding for the final result.

Recently, end-to-end speech recognition systems have become increasingly popular and achieve promising performance in mandarin \cite{amodei2016deep}. 
End-to-end speech recognition methods predict graphemes directly from the acoustic data without linguistic knowledge, thus reducing the effort of building ASR systems greatly and making it easier for new language.
The end-to-end ASR simplifies the system into a single network architecture, and it is likely to be more robust than a multi-module architecture.
There are two major types of end-to-end architectures for ASR: 
The connectionist temporal classification (CTC) criterion \cite{graves2006connectionist,kim2017joint,chiu2017state,graves2014towards}, 
which has been used to train end-to-end systems that can directly predict grapheme sequences.
The other is attention-based encoder-decoder model \cite{prabhavalkar2017comparison,bahdanau2016end,chorowski2016towards,chan2016online} which applies an attention mechanism to perform alignment between acoustic frames and recognized symbols.

Attention-based encoder-decoder models have become increasingly popular \cite{shan2017attentions,chiu2017state,chorowski2015attention,hori2017advances}. These models consist of an encoder network, which maps the input acoustic sequence into a higher-level representation, and an attention-based decoder that predicts the next output symbol conditioned on the full sequence of previous predictions.

A recent comparison of sequence-to-sequence models for speech recognition \cite{prabhavalkar2017comparison} has shown that Listen, Attend and Spell (LAS) \cite{LAS}, a typical attention-based approach, offered improvements over other sequence-to-sequence models, and attention-based encoder-decoder model performs considerably well in mandarin speech recognition \cite{shanattention}. 

For Mandarin speech recognition, modeling units of acoustic model affect the performance significantly \cite{chang2000large}. As we all know, CDP is most commonly used as the acoustic modeling units for speech recognition in mandarin \cite{amodei2016deep}. In fact, there have been several different kinds of modeling units for Mandarin \cite{xiangang2015comparative} such as phoneme, syllable and Chinese character. Compared with CDP, it will be easier to use syllable or character which does not need other prior model for alignment. Under current end-to-end speech recognition framework, we can get target output syllable sequence and character sequence directly from training transcripts and lexicon. Especially, in the case of using Chinese character models, we can get the desired results directly without lexicon and language model. 

In order to find a more suitable end-to-end system and modeling unit in Mandarin speech recognition, we explore two major end-to-end models: CTC model and attention based encoder-decoder model. Meanwhile, We compare the performance of three different scaled modeling units: context dependent phoneme (CDP) , syllable with tone and Chinese character.

The rest of this paper is organized as follows. 
Section \ref{section_end-to-end} introduces the details of end-to-end speech recognition. 
Various model units for end-to-end speech recognition in mandarin are studied in Section \ref{section_model_untis}.
Section \ref{section_exp} describes the detail of the experiments.
Section \ref{section_conclusion} draws some conclusions and outlines our future work.

\section{End-to-End Speech Recognition}\label{section_end-to-end}
Recently, end-to-end speech recognition systems have become increasingly popular and achieve encouraging performance in mandarin. 

\subsection{Connectionist Temporal Classification(CTC)}\label{section_CTC}
The CTC criterion was proposed by Graves et al. \cite{graves2006connectionist} as a way of training end-to-end models without requiring a frame-level alignment of the target labels for a training utterance.
To achieve this, an extra ‘blank’ label denoted $\langle b \rangle$ is introduced to map frames and labels to the same length, which can be interpreted as no target label.
CTC computes the conditional probability by marginalizing all possible alignments and assuming conditional independence between output predictions at different time steps given aligned inputs.

Given a label sequence $y$ corresponding to the utterance $x$, where $y$ is typically much shorter than the $x$ in speech recognition.
Let $\beta(y, x)$ be the set of all sequences consisting of the labels in $\mathcal{Y} \cup {\langle b \rangle}$, which are of length $|x| = T$, and which are identical to $y$ after first collapsing consecutive repeated targets and then removing any blank symbols (e.g., $A{\langle b\rangle}AA{\langle b \rangle}B \to AAB$).
CTC model defines the probability of the label sequence conditioned on the acoustics as Equation \ref{CTC_prob}.

\begin{equation}
  P_{CTC}(\mathbf{y}|\mathbf{x}) = \sum_{\hat{\mathbf{y}} = \beta(\mathbf{y},\mathbf{x})} P(\hat{\mathbf{y}}|\mathbf{x})
                                = \sum_{\hat{\mathbf{y}}=\beta(\mathbf{y},\mathbf{x})} \prod_{t=1}^{T} P(\hat{y}_t|\mathbf{x}) \label{CTC_prob}
\end{equation}

With the conditional independent assumption, $P_{CTC}(\hat{\mathbf{y}}|\mathbf{x})$ can be decomposed into a product of posterior $P(\hat{y}_t|\mathbf{x})$ in each frame $t$. The conditional proability of the labels at each frame, $P_{CTC}(\hat{y}_t|\mathbf{x})$, can be estimated using BLSTM, which we refer to as the encoder. The model can be trained to maximize Equation \ref{CTC_prob} by using gradient descent, where the required gradients can be computed using the forward-backward algorithm \cite{graves2006connectionist}.

CTC models have a conditional independence assumption on its outputs, wherein it will become difficult to model the interdependencies between words.
During the beam search process, language model and word count are introduced.
The beam search process of CTC \cite{wiseman2016sequence} is to find 
\begin{equation}
  {\arg\max}_{y} ({\log(P_{CTC}(y|x)) + \alpha \log(P_{LM}(y)) + \beta wordcount(y)})  \label{CTC_search}
\end{equation}
where a language model and word count are included, and $\alpha$ and $\beta$ are the weights of them respectively.

\subsection{Attention based models}
Chan et al. \cite{LAS} proposed Listen, Attend and Spell (LAS), a kind of neural network that learns to transcribe speech utterances to characters.
As an attention-based encoder-decoder network, LAS is often used to deal with variable length input and output sequences.
Using the attention mechanism, the attention model can align the input and output sequence.

As section \ref{section_CTC} mentioned, the CTC assumes monotonic alignment, and it explicitly marginalizes over alignments. 
And because of the conditional independence assumption, the CTC model can not explicitly learn co-articulation patterns, which exist in speech commonly.
Attention based models remove the conditional independence assumption in the label sequence that CTC requires, then the $p(y|x)$ defines as Equation \ref{Attn_prob}

\begin{equation}
   P_{Attention}(\mathbf{y}|\mathbf{x}) =  P(\mathbf{y}|\mathbf{h}) = \prod_{t=1}^{T} P(y_t|c_t, y_{<t}) \label{Attn_prob}
\end{equation}
where $c_t$ is the context at decoding time step $t$.

An attention-based model contains an encoder network and an attention based decoder network.
The encoder network maps the input acoustics into a higher-level representation.
The attention based decoder network predicts the next output symbol conditioned on the full sequence of previous predictions and acoustics, which can be defined as $P(y_u|y_{u-1},\cdots, y_1, x)$.
The attention mechanism selects or weights the input frames to generate the next output label.

\begin{figure}[t]
  \centering
  \includegraphics[width=6cm,height=6cm]{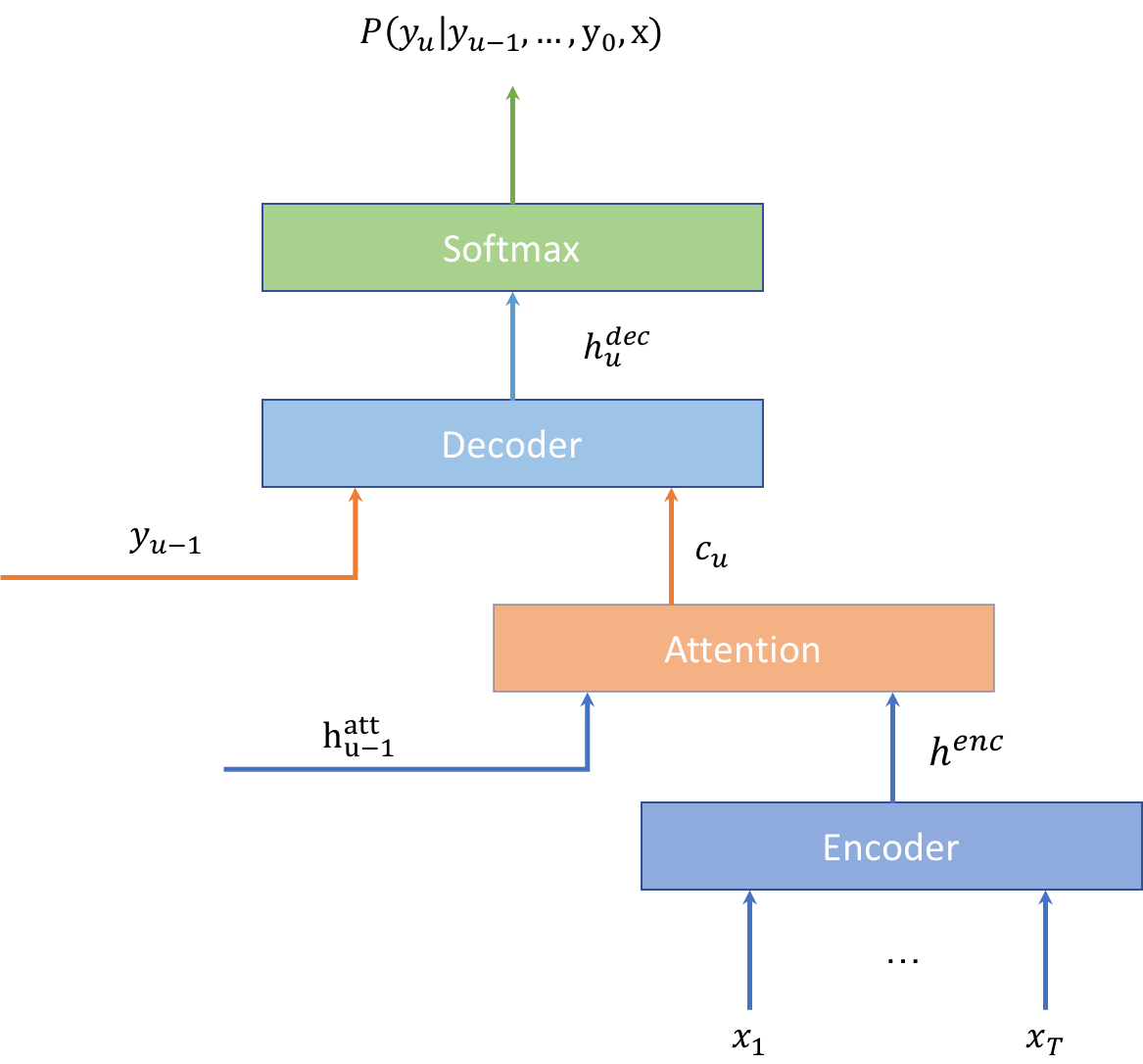} 
  \caption{Schematic diagram of the attention-based encoder-decoder network.}
  \label{fig:attention}
\end{figure}

As shown in Figure~\ref{fig:attention}, the attention-based encoder-decoder network can be defined as:
\begin{equation}
  h = Encoder(x)  \label{encoder}
\end{equation}
\begin{equation}
  P(y_u|x,y_{1:t-1})= AttentionDecoder(h, y_{1:t-1}) \label{decoder}
\end{equation}
where $Encoder(\cdot)$ can be long short-term memory(LSTM) or bidirectional LSTM (BLSTM) and $AttentionDecoder(\cdot)$ can be LSTM or gated recurrent unit(GRU).

The beam search process of attention is to find 
\begin{equation}
  {\arg\max}_{y} ({\log(P_{Att}(y|x)) / |y|^{\gamma} + \beta cov(\alpha) +  \lambda \log(P_{LM}(y))})  \label{Attention_search}
\end{equation}
where $\gamma$ is the length normalization hyperparameter. The coverage term "cov" encourages the model to attend over all encoder time steps, and stops rewarding repeated attendance over the same time steps. The coverage term addresses both short as well as infinitely long decoding.

\section{Acoustic Modeling Units}\label{section_model_untis}
\begin{figure}[t]
  \centering
  \includegraphics[width=\linewidth]{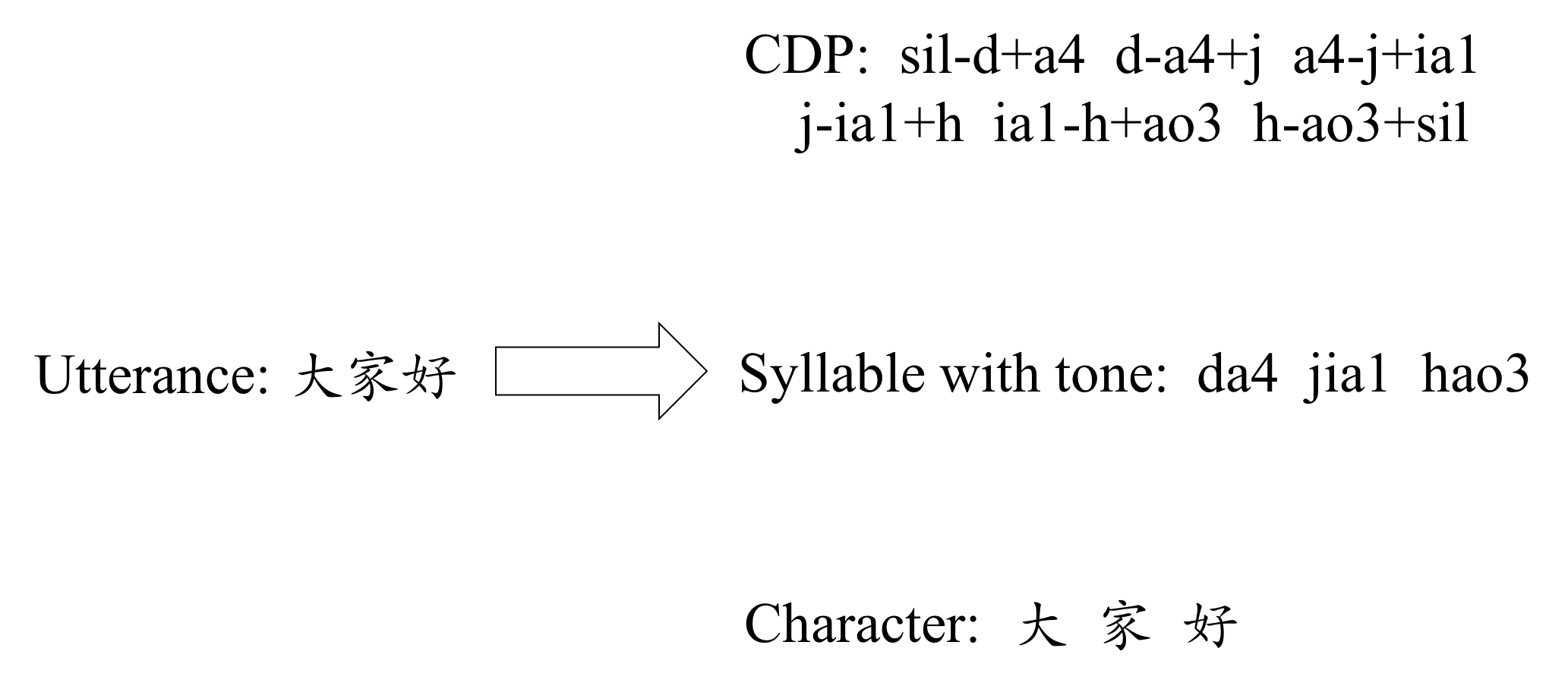}
  \caption{An example of converting one Chinese utterance into various modeling units.}
  \label{fig:model_unit_example}
\end{figure}

In mandarin speech recognition, modeling units of acoustic model affect the performance significantly.
There have been kinds of different acoustic representations for Mandarin in recent years \cite{chang2000large,xiangang2015comparative,li2004context}. For example,
There have been syllable initial/final approach, syllable initial/final with tone approach, syllable approach, syllable with tone approach, Chinese Character approach and preme/toneme approach \cite{chen1997new}. 
In this study, we select context dependent syllable initial/final with tone, syllable with tone and Chinese Character as study object.
Figure~\ref{fig:model_unit_example} shows an example of various modeling units.

\subsection{Context Dependent Phoneme (CDP)}
For CTC based end-to-end speech recognition in mandarin, CDP is commonly used as the acoustic modeling unit. 
We usually use syllable initial/final with tone as phoneme, such as syllable initial $d$ and syllable final with tone $a4$,  and the context dependent phoneme is like $sil$-$d$+$a4$.

\subsection{Syllable}
A syllable with tone consists of a syllable initial and a syllable finial with tone,\ such as $da4$. Chinese is naturally a syllabic language and each basic language unit (Chinese character) can be phonetically represented by a syllable \cite{wu2007context}. Furthermore, each Chinese syllable also has syllable Initial-Final structure. According to the official released scheme for Chinese phonetic alphabet, each syllable is regarded as the
combination of these aspects are very helpful for the design of acoustic models.

\subsection{Character}
Like English words, Chinese characters are the basic symbols of the recording language. In most cases, in mandarin speech recognition, our goal is to transcribe the speech sequence into the Chinese Character sequence. Therefore, in the end-to-end speech recognition framework, Chinese character is a perfect modeling unit which can be decoded without language model and lexicon.

There is no exact number of Chinese characters, the number is about one hundred thousand, and there are only a few thousand characters in the daily use of Chinese characters. In our work, We chose 4977 common Chinese characters and the coverage is 99.92\% on our datasets. 

\section{Experiments}\label{section_exp}
Several experiments have been done to compare the performance of three kinds of acoustic modeling units by the two types of end-to-end methods. 
We find that both On the DidiReading dataset and DidiCallcenter dataset the Character based attention models achieve the best performance.
\subsection{Data}
We do all the experiments both on DidiCallcenter dataset and DidiReading dataset, which are different not only on the the data size but also the dialogue scene. 
The DidiCallcenter dataset contains more than 2.2M utterances (about 2,800 hours), which is a spontaneous style dataset.
The DidiReading dataset contains more than 16.2M utterances (about 12,000 hours), which is a reading style dataset.
There are also two test sets, which are randomly extracted from the two datasets respectively. The DidiCallcenter test set includes 2000 utterances and the DidiReading test set includes 5000 utterances.
40 mel-scale filter-banks coefficients computed every 10ms are used as input features for both datasets.
Global mean and variance normalization is conducted for each dataset.

Table~\ref{tab:labels} shows the detailed information of the labels for various modeling units.

\begin{table}[]
\centering
\caption{Detailed composition of various labels}
\label{tab:labels}
\begin{tabular}{lll}
Models    &Modeling Units &Composition of the label \\
\midrule
CTC       & CDP            &DidiCallcenter: $\sim$ 12,100CDP\\
          &                &                + 1 BLANK$\langle b \rangle$ \\
          &                &DidiReading: $\sim$ 12,200CDP\\
          &                &            + 1 BLANK$\langle b \rangle$ \\
          & Syllable       &1313 tonal syllable with tone\\
          &                &                     + 1 BLANK$\langle b \rangle$\\   
          & Character      &4977 Chinese character\\
          &                &{              + 1 BLANK$\langle b \rangle$}\\       
\midrule
Attention & Syllable       &1313 tonal syllable with tone\\
          &                &            + 1 unknown token$\langle unk \rangle$\\
          &                &     + 1 sentence start token$\langle sos \rangle$\\
          &                &+ 1 sentence end token$\langle eos \rangle$\\
          & Character      & 4977 Chinese characters\\
          &                &            + 1 unknown token$\langle unk \rangle$\\
          &                &     + 1 sentence start token$\langle sos \rangle$\\
          &                &+ 1 sentence end token$\langle eos \rangle$\\     
    \bottomrule

\end{tabular}
\end{table}

\subsection{CTC models}
In this work, CTC models are trained to predict CDP, syllable and character as output targets, respectively.

\subsubsection{Training}
\begin{figure}[t]
  \centering
  \includegraphics[width=3cm,height=7.5cm]{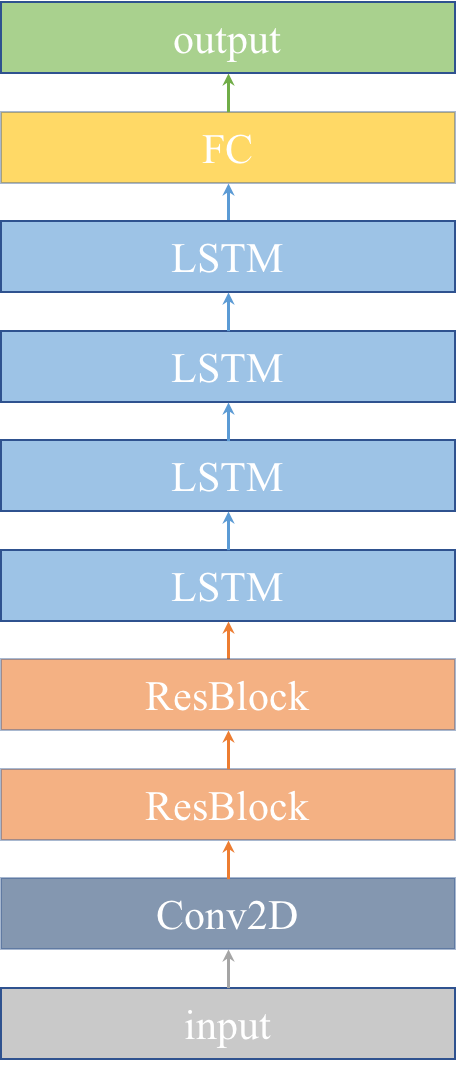}
  \caption{The network architectures of CTC model.}
  \label{fig:res-ctc}
\end{figure}

The network architecture of CTC is described in Figure~\ref{fig:res-ctc}, which contains one convolutional-2D layer, two residual blocks \cite{zhang2017very}, four LSTM \cite{sainath2015convolutional} layers and one full-connection layer.
Each residual block includes two convolutional-2D layers. 
Each LSTM layer contains 1024 nodes and followed by layer normalization. 
The parameters of CDP-CTC, Syllable-CTC and Character-CTC model are about 86M, 30M, 46M, respectively. 
During training stage, Adam \cite{kingma2014adam} optimization method is used and L2 weight decay is 1e-5, the learning rate is decayed from 1e-3 to 1e-6 during training.

\subsubsection{Decoding}
These models are decoded using external 4-gram Chinese word language models. For DidiCallcenter task, the size of language model is 40GB which contains 1.9G gram tokens, on the other DidiReading task, we use a 55GB language model which contains 2.7G gram tokens.

\subsection{Attention models}
In this work, attention models are trained to predict syllable and Character as output targets, respectively.

\subsubsection{Training}
For syllable attention experiments, our models are LAS models with 2 convolutional layers, followed by 4 bi-directional LSTM layers with 256 LSTM units per-direction, interleaved with 3 time-pooling layers which resulted in an 8-fold reduction of the input sequence length. The Decoder was a 1 layer LSTM with 256 LSTM units and output has 1316 labels. For Character experiments, our LAS models has the same architecture as the Syllable model, except that the output has 4980 labels. The syllable attention model has about 8.79M parameters and character attention model has about 12.54M parameters.
 During training stage, schedule sampling and unigram label smoothing is applied as described in \cite{chiu2017state,bahdanau2016end,chorowski2016towards}, Adam optimization method with gradient clipping is used for optimization. We initialized all the weights randomly from an isotropic Gaussian distribution with variance 0.1 and learning rate is decayed from 5e-4 to 5e-6 during training. All models are trained with the cross-entropy criterion and are trained using TensorFlow \cite{abadi2016tensorflow}.

\subsubsection{Decoding}
A left-to-right beam search over modeling unit sequences was used during decoding. Beam search was stopped when the sentence end token  $\langle eos \rangle$  was emitted. We also integrated external language models during decode stage and all the language models were trained with the training transcripts.

\subsection{Results}
We first conduct experiments on different modeling unit for CTC model. From Table~\ref{tab:ctc-model-unit}, we can find that all modeling units can achieve similar CER but syllable based CTC model achieves the best performance both on the DidiCallcenter dataset and DidiReading dataset. Meanwhile, because syllable-based model has much less parameters than CDP-based model, the time model needed to converge is much less and decoding is much faster. Therefore, we believe that syllable is a more suitable modeling unit for CTC model.

\begin{table}[th]
  \caption{CER(\%) of CTC-based method on various modeling units (\#Param is the number of model parameters)}
  \label{tab:ctc-model-unit}
  \centering
  \begin{tabular}{c c c c}
    \toprule
    \multicolumn{1}{c}{\textbf{Models}} & 
    \multicolumn{1}{c}{\textbf{\#Param}} &    
    \multicolumn{1}{c}{\textbf{Callcenter}} &
    \multicolumn{1}{c}{\textbf{Reading}} \\
    \midrule
    \midrule
    \multirow{1}{2cm}{CDP-CTC} & 86.05M  & 7.42 & 5.81 \\
    \multirow{1}{2cm}{Syllable-CTC} & 30.04M & \bf{7.31} & \bf{5.62} \\
    \multirow{1}{2cm}{Character-CTC} & 46.10M& 7.45 & 5.79 \\
    \midrule                       
    \bottomrule
  \end{tabular}
\end{table}

Then, we compare the results of attention-based models in Table~\ref{tab:attention-model-unit}. By comparing the performance of the syllable attention model and character attention model, it's clear to see that the performance of character-model is better than syllable-model. We believe it's because the language model we're using isn't strong enough and the implicit RNN language model decoder learned helps to boost the performance of character-model. At the same time, we find external language model to be helpful for both of our tasks. But DidiReading task benefits more from external language model, we think it's because the DidiReading task is a is domain-specific task which contains a lot of special terms, and the external language model can help solve this problem effectively.

\begin{table}[th]
  \caption{CER(\%) of Attention-based method on various modeling units(\#Param is the number of model parameters)}
  \label{tab:attention-model-unit}
  \centering
  \begin{tabular}{c@{} c c c c}
    \toprule
    \multicolumn{1}{c}{\textbf{Models}} & 
    \multicolumn{1}{c}{\textbf{Units}} &
    \multicolumn{1}{c}{\textbf{\#Param}} &    
    \multicolumn{1}{c}{\textbf{Callcenter}} &
    \multicolumn{1}{c}{\textbf{Reading}} \\
    \midrule
    \midrule
    \multirow{1}{1.7cm}{Attention}  & Syllable & 8.79M & - & - \\
    \multirow{1}{1.7cm}{   + LM}  & Syllable & 8.79M &6.34  &5.78  \\
    \midrule
    \multirow{1}{1.7cm}{Attention}  & Character & 12.54M &5.86 &6.22  \\  
    \multirow{1}{1.7cm}{   + LM}  & Character & 12.54M &\bf{5.68}  &\bf{4.89}  \\
    \midrule
    \bottomrule
  \end{tabular}
\end{table}

A comparison of the CTC models and attention models reveals some interesting conclusions.
First, we note that the performance of attention model is significantly better than CTC model. On the DidiCallcenter task, the CTC model achieves a best CER of 7.31\%, by using the character attention model, we improved the CER to 5.68\%. In the same way, we improved the CER from 5.62\% to 4.89\% on the DidiReading task. It is also interesting to compare the structure of the two types of models, our CTC models used un-directional LSTM and attention models used bi-directional LSTM, but the number of parameters in CTC models is three times as much as attention models. In future work, we hope to compare a bigger attention model and bi-directional LSTM CTC model which has the same number of parameter as attention model.

\section{Conclusions}\label{section_conclusion}
In this work, we studied the performance of various acoustic modeling units on different end-to-end models in Mandarin speech recognition.
 
In experimental evaluations, we find that on CTC model, syllable achieve the best CER on both DidiCallcenter dataset and DidiReading dataset.
Moreover, we find that all modeling units can achieve approximate CER in CTC model.
On attention model, however, we find that character model outperforms syllable model. 
Namely, Chinese character is a appropriate modeling unit for acoustic modeling.
Finally, we compare two end-to-end models and find that attention model is much better than CTC model in Mandarin speech recognition, even if the size of the attention model is much smaller than CTC model.

\section{Acknowledgements}
The authors would like to thank Liqiang He, Caixia Gong, Xiaohui Li and Mingxin Liang for their help.  

\bibliographystyle{IEEEtran}

\bibliography{model-unit-asr}

\end{document}